# A system for generating complex physically accurate sensor images for automotive applications

Zhenyi Liu[1,2], Minghao Shen[2], Jiaqi Zhang[3], Shuangting Liu[3], Henryk Blasinski[1], Trisha Lian[1], Brian Wandell[1]
1. Stanford University, 2. Jilin University, 3. Beihang University

## Abstract

*We describe an open-source simulator that creates sensor irradiance and sensor images of typical automotive scenes in urban settings. The purpose of the system is to support camera design and testing for automotive applications. The user can specify scene parameters (e.g., scene type, road type, traffic density, time of day) to assemble a large number of random scenes from graphics assets stored in a database. The sensor irradiance is generated using quantitative computer graphics methods, and the sensor images are created using image systems sensor simulation. The synthetic sensor images have pixel level annotations; hence, they can be used to train and evaluate neural networks for imaging tasks, such as object detection and classification. The end-to-end simulation system supports quantitative assessment – from scene to camera to network accuracy - for automotive applications.*

## Introduction

The massive growth of the mobile imaging market produced enormous innovation in camera design. Cameras for consumer photography are designed around image quality metrics that account for human observers (Wang et al. 2004). The expected massive growth in imaging applications for machine-learning (ML) motivates us to consider the potential for new camera designs. Camera metrics for ML applications should include metrics that address the critical requirements of ML: accuracy and generalization. Specifically, we must understand how well networks trained using images from one camera perform with inputs from a different camera.

Classically, domain adaptation methods are used to adjust for the differences between the available data and the target domain (Wu and Dietterich 2004; Duan et al. 2009). Two recent techniques specifically address the limits of generalization between synthetic and real images. Domain randomization introduces random variations into the synthetic image with the hope that such perturbations force the network to focus on critical information (Tremblay et al. 2018). Domain stylization uses photorealistic image style transfer algorithms to transform synthetic images so that an independent network cannot discriminate synthetic and measured images (Dundar et al. 2018; Li et al. 2018). There is also valuable ongoing work to improve the realism of synthetic images, which can be automatically labeled and used for training and ML validation (Tsirikoglou et al. 2017; Wrenninge and Unger 2018).

A limitation of domain adaptation methods for our purpose is that they do not explicitly represent camera parameters. Hence, this limits the ability to explore optics and sensor factors such as the lenses, color filters, or pixel size. For guiding camera design, it is essential to incorporate real camera parameters when creating synthetic images. Because of this requirement, the simulation must be able to create physically accurate synthetic scene radiance distributions that enable us to model the impact of wavelength-dependent components, including the optics and sensors (Blasinski et al. 2018).

This paper describes an open-source and freely distributed toolbox to synthesize scene spectral radiances and sensor data for neural network automotive applications. The software includes procedural methods to generate a large number and variety of scenes from graphics assets stored in a database. The software simulates optics and sensors to support the exploration of novel camera designs. We are using these synthetic sets with ML systems to develop metrics that evaluate the performance and generalization of camera and ML systems.

## Methods

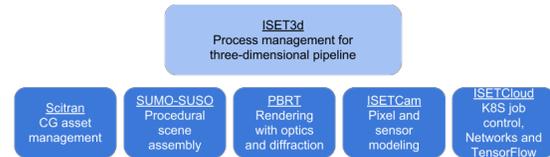

**Figure 1.** ISET3d scripts control and set parameters for multiple software components. These components randomly sample computer graphics (CG) assets from the Flywheel database (Scitran), assemble the data into a scene spectral radiance (SUMO-SUSO) that is rendered into a spectral irradiance at the sensor surface (PBRT) as well as pixel level metadata (depth, object, and material labels). The irradiance is transformed into a sensor response based on a pixel and sensor model (ISETCam). The collection of images is grouped into data sets that are used to train and evaluate networks (ISETCloud).

The simulation tools combine several components (Figure 1). Users control the overall computation, from scene generation to sensor simulation, using scripts in the Matlab toolbox ISET3d. The functions in this toolbox control the critical stages of collecting and assembling the scene assets, modeling the scene parameters (e.g., surface properties, object and camera positions), calculating the scene radiance, modeling the sensor properties and calculating the sensor response, and invoking the ML application.

Computer graphics assets are stored in a database that is accessed using the Matlab SciTran toolbox. This toolbox addresses the Flywheel.io database to randomly sample and download objects and sky maps. The density, positions, and velocity of the objects (cars, pedestrians, cyclists, signs and traffic lights, trees) are defined by an open source microscopic road traffic simulator (Simulation of Urban Mobility, SUMO)(Behrisch et al. 2011). We also developed software (Simulation of Urban Static Objects, SUSO) to calculate the positions of static objects (buildings). The collection of objects and their properties is assembled into two files. The recipe file, stored in JSON (Javascript object notation) format, lists the objects and their parameters. The resources file includes the data necessary for rendering the objects (e.g., textures in png format and meshes (points, triangle vertices, faces) in pbrt format).

PBRT, an open-source, physically based ray-tracing software, is the critical component that implements ray tracing (Pharr, Jakob, and Humphreys 2016). PBRT uses principles of physics to model the interaction between light and matter in the 3D world. PBRT simulates physically accurate spectral radiance, transforming the radiance into sensor irradiance by implementing a camera lens model. We added features to PBRT to enable specification of a multi-element camera lens comprised of spherical, aspherical, or biconic surfaces with arbitrary wavelength-dependent indices of refraction and to account for the effects of diffraction (Freniere, Groot Gregory, and Hassler 1999). Finally, we inserted methods so that PBRT generates images comprising the metadata used for ML training (distance maps, object labels, material labels). To simplify sharing, we placed the modified PBRT into a Docker Container that includes its dependencies. The container runs on a wide variety of computational platforms without the need for compilation.

ISETCam converts spectral irradiance data into pixel responses. This code enables the user to specify a large range of pixel and sensor properties and it also includes methods for the image systems pipeline (Farrell, Catrysse, and Wandell 2012; Farrell and Wandell 2015). Finally, ISETCloud manages the cloud-scaling of rendering jobs via Kubernetes running on the Google Cloud Platform (GCP). This software also sends training and evaluation jobs that run using Tensorflow on the GCP. The collection of software tools described here significantly extend the system described in a previous paper (Blasinski et al. 2018).

## Results

### Accumulation of assets

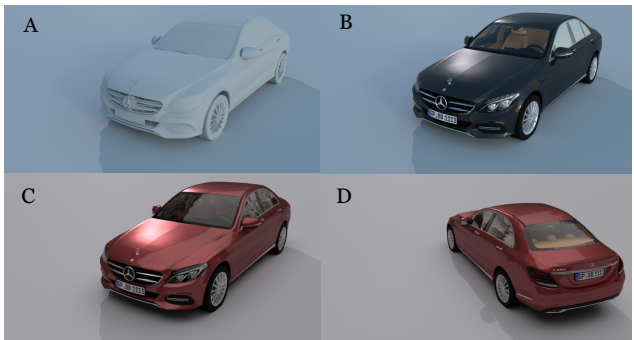

**Figure 2**. (A) The mesh defining a car is rendered as a Lambertian surface. (B) Different materials are assigned to the various car parts. (C) Different material color and new lighting are assigned. (D) The car is rotated but the camera position is unchanged.

Our database currently contains computer graphics assets representing more than 200 buildings, 100 vehicles, 80 pedestrians, and other street elements (e.g., trees, signs, traffic lights). Each asset is stored separately with its own 3d description file (assetRecipe.json) and resources file (pbrt format) for rendering. Figure 2 illustrates a typical car asset, rendered with different types of surface material, color, and lighting.

This growing collection of assets were obtained by converting open-source assets from various sources (e.g., Blender, Maya, Adobe, commercial vendors). The parts of these open assets (e.g., doors, tires, windshield) were not consistently labeled, and thus we edited the labels and scaled all the sizes to meters. Most of the assets were converted to the PBRT asset format using a utility that converts Cinema4D data into the PBRT format that is part of the PBRT distribution.

### Scene assembly

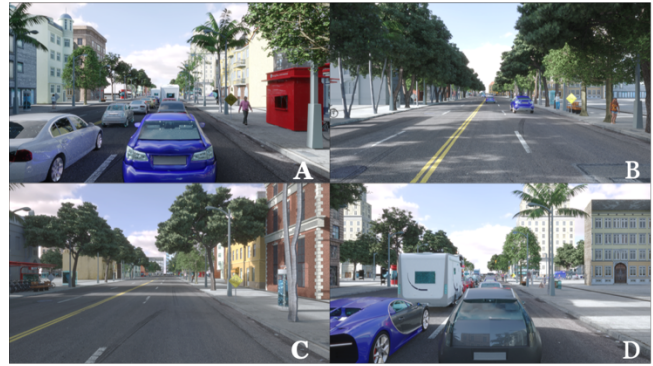

**Figure 3.** Scenes generated on the same road and sky map, but randomly sampling buildings, trees, environmental objects and setting different traffic and environmental parameters. (A,D) High traffic density. (B,C) Low traffic density. The camera position is chosen to be consistent with the car that is acquiring the image (not shown).

Synthetic scenes for ML applications should be able to vary, matching the complexity of the real world (Figure 3). To manage a large number of assets needed to create realistic scenes we find it useful to store the assets and their metadata (e.g., high level verbal descriptions) in a searchable database (MongoDB). We use a database (https://flywheel.io) that runs on the Google Cloud Platform (GCP) and can be accessed from any computer on the Internet. The database also has a software development kit that enables the user to query the database in a number of programming languages and to download collections of assets.

The software includes functions that place the mobile assets (cars, people, buses, trucks, bicyclist) into a realistic spatial organization on the streets using Simulation of Urban MObility (SUMO) (Behrisch et al. 2011). Users can specify scenes either by naming specific assets or by setting statistical parameters to control random selection methods. This procedure creates quite complex and realistic traffic scenarios, including scenes with pedestrians crossing roads, vehicles changing lanes, vehicle braking when the traffic light turns red, etc. To add further variation to each rendered scene, we developed software to insert static assets, Simulation of Urban Static Objects (SUSO). This software randomly samples static assets from the database (buildings, trees, street lights, etc.,) and places them into the scene.

The collection of assets and asset positions form an important part of the scene 'recipe'. This is a JSON file that is built up during the simulation process. The file includes the information about the scene metadata. The assets needed for the rendering are collected up into a zip-file of computer graphics resources (cgresources.zip).

### Scene rendering

Once assets are assembled and placed, we use PBRT to calculate the scene spectral irradiance at the sensor surface. The calculation follows the conventional ray tracing approach: multiple rays are sent from each point on the sensor surface through the optics into the scene.

We use the extensibility of the PBRT software to add several computational modules, including extended lens modeling as described in the Methods section. In addition, we added a module to randomize the directions of rays passing near an aperture (Freniere, Groot Gregory, and Hassler 1999). We can use different lens description files to simulate different lens effect (Figure 4A).

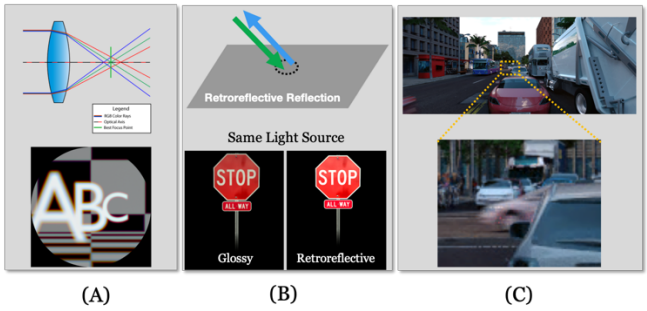

**Figure 4.** (A) Optics: Chromatic Aberration; (B) Material: Retroreflective; (C) Non-uniform Motion blur

Also, we added a retroreflective material class to model the properties of certain street signs (Figure 4B). In this case, rays incident on the surface are likely to be reflected back in the direction from which they arrive. PBRT also incorporates rendering methods that calculate the impact on the irradiance when certain assets have linear motion (Figure 4C). In PBRT, a transformation matrix is used to move an object from one point to a different point through translation, rotation and scale. When rendering motion blur in the system, we define the camera shutter rate, start and end transformation matrices for certain objects, and a transformation time (one second by default). Thus, cars traveling in different directions at different speeds(meters/second) can be rendered accurately. In addition, the PBRT realistic camera model can simulate different types of lenses (Figure 5). A number of different lens models are included in the software distribution, and more can be added if the user has lens prescription data are available.

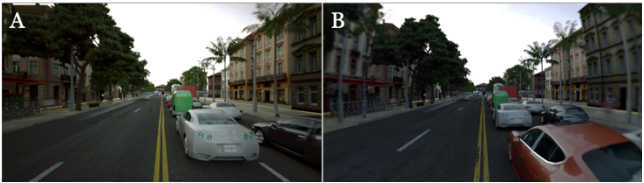

**Figure 5**. A comparison of cameras with different imaging lenses. (A): wide angle lens; (B): fisheye lens

Real-world datasets require labeling, either manually or by AI algorithms. In either case, quality assurance is necessary to reduce errors. To obtain depth information, an extra device (lidar) is usually needed. However, for synthetic PBRT renderings, a ray from the camera to the scene can return object information, including shape, surface properties, and position. We implemented modifications to PBRT to return these labels for depth map, class, and instance label (Figure 6). Hence, accurate pixel-level labels can be generated at a relatively low computational cost.

### Sensor modeling

The *ISETCam* [Ref. 8] software converts the spectral irradiance data into pixel voltages and digital outputs (Farrell, Catrysse, and Wandell 2012; Farrell and Wandell 2015). Using ISETCam we can simulate different types of sensors, varying parameters such as the color filter array and pixel size (Figure 7).

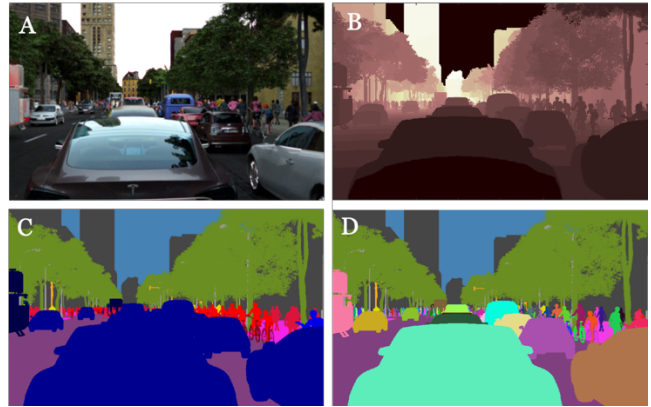

**Figure 6.** The PBRT simulation produces sensor irradiance images as well as metadata. (A) The irradiance data can be converted into a camera sensor image; (B-D) The PBRT simulation also produces pixel-level metadata of depth, object class, and specific object instance. The metadata are necessary for machine-learning applications.

We were particularly interested to understand the effects of pixel size on the ability to identify cars and pedestrians in a typical driving scene. Thus, we simulated sensors that were matched in all ways with the exception that Sensor A has a 6-micron pixel and Sensor B has a 3-micron pixel. Maintaining the field of view means that Sensor A is 752 x 480, while sensor B is 1504 x 960. All other parameters (color filters, frame rate, dark noise, read noise, PRNU, DSNU, and exposure duration) were equated.

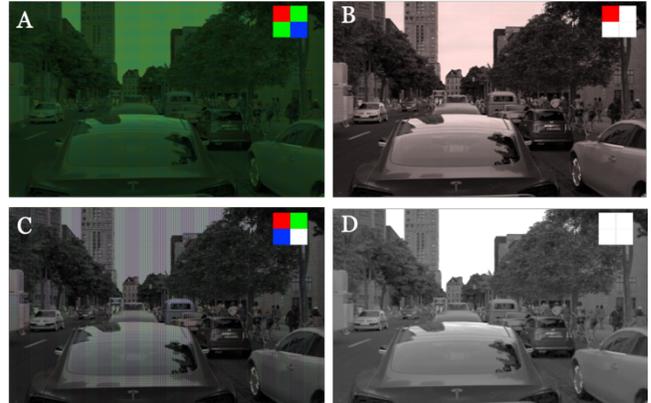

**Figure 7**. The sensor irradiance data can be used to calculate sensor responses for arbitrary color filter arrays. The four images illustrate sensor responses for four different types:(A) RGGB (B) RCCC (C) RGBW, and (D) Monochrome.

### *Evaluating the impact of alternative camera designs*

Changes to most camera parameters involve a trade-off in functionality. For example, shrinking pixel size increases the sampling resolution of the image and reduces signal-to-noise at each pixel. A critical goal of these simulations is to produce enough synthetic data that we can evaluate these tradeoffs by quantifying network performance. This should help us make informed camera design decisions.

As an illustration, we evaluated the performance of two identical cameras, differing only in pixel size (3-micron vs. 6-micron) and thus the number of pixel samples (see Appendix I for the sensor parameters). We fine-tuned a coco pre-trained Tensor Flow Models [Ref. 10], Faster RCNN Resnet101 to identify cars and pedestrians from a sample of synthetic images estimated from these two cameras. The network was fine-tuned on 8000 images from Berkeley Deep Drive Dataset (Yu et al. 2018) and then tested on 1500 held-out synthetic images that we created. The test set includes the images we generated with random parameters for traffic flow simulation, we think the large test set of randomly generated traffic scenes is representative of the real-world driving scenarios. We measured the average precision for recognizing cars and pedestrians in the test set as a function of distance, a quantity that depends on the sensor spatial resolution (Figure 8).

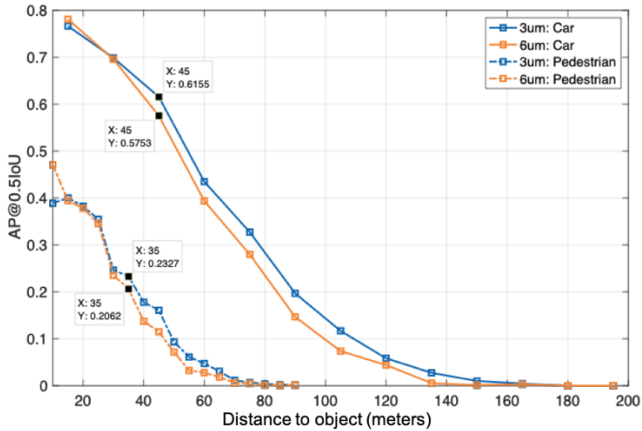

**Figure 8**. Sensor parameters for the simulation in Tables 1 and 2 (Appendix).

The curves for both cars and pedestrians follow a similar pattern: average precision declines over distance. The performance of the 6-micron camera matches the 3-micron camera until a distance of about 40 meters, and then it is slightly less accurate from 40-150 meters. Under the simulated day time driving conditions, the system with the smaller pixel sees about 5 meters further at equal performance level. It is likely that under low light conditions there will be benefits from the 6-micron camera, although we have not run that particular simulation

We use this example to demonstrate the ability to make specific performance evaluations of a meaningful automotive task. The evaluation includes factors spanning the scene, the camera, and the network. Camera design can best be supported by the ability to evaluate the tradeoffs and help us improve the full imaging system - from scene to camera to algorithm.

## Discussion

A principal goal of this paper is to create a platform that will enable us to experiment with camera and sensor designs. The key novelty of this work is the ability to synthesize a large number of scene spectral radiance data that can be used as inputs to simulated cameras. A critical advance compared to our previous work is the insertion of the SUMO-SUSO software; these tools create large numbers of scenes from the Flywheel database of assets. The ability to create scenes using an automated procedure enables us to sweep out a statistical specification of the driving conditions (e.g., traffic flow density of different assets) and create many possible scenes.

There were additional advances in the PBRT rendering methods as well, so that the synthetic spectral irradiance data are increasingly accurate. Spectral irradiance data, rather than synthetic RGB images, are necessary to calculate the output of experimental camera designs.

Finally, these tools enable us to create new metrics that evaluate the tradeoff arising from selecting different camera parameters. Assessing the tradeoffs in ML performance for object detection as a function of targets at different distances is one performance metric, just as changing pixel size is one camera design example. There is an opportunity to design many metrics that may be helpful, and we hope to pursue this direction in future work.

## Appendix

| Parameter | Specification A | Specification B |
|---|---|---|
| Optical Format | 1/3 inch | 1/3 inch |
| Active Imager Size | 4.51 mm(H) x 2.88 mm(V) | 4.51 mm(H) x 2.88 mm(V) |
| Active Pixels | 752H x 480 V | 1504H x 960 V |
| Pixel Size | 6 um x 6 um | 3 um x 3 um |
| Color Filter Array | RGGB Pattern | RGGB Pattern |
| Full Resolution | 752 x 480 | 1504 x 960 |
| Frame Rate | 60fps | 60fps |
| Dark Noise | 1.0mV/pixel/second | 1.0mV/pixel/second |
| Read Noise | 1.0 mV | 1.0 mV |

**Table 1**. Sensor parameters for the evaluation results shown at Figure 8. In both cases, the optics were a wide angle, 112-degree field of view and a 6 mm focal length.

| Parameter | Specification |
|---|---|
| Feature Extractor | faster_rcnn_resnet101 |
| Number of Layers | 101 |
| First stage features stride | 16 |
| First stage IOU threshold | 0.7 |
| First stage max proposals | 300 |
| Optimizer | momentum optimizer |
| Regularizer | l2_regularizer |
| Score converter | SOFTMAX |

**Table 2**. Network model parameters ("Faster RCNN Resnet 101"). The model was run on the GCP using TensorFlow.

## Acknowledgements

We thank Dr. Joyce Farrell for her contributions to this work. The project was funded by the Recruitment Program of Foreign Experts, 1000 Talents Program, Jilin University (ZL, HB, TL, BW) and summer student funding from the Chinese government (MS, JZ, SL). We thank Professor Weiwen Deng, Beihang University, for his advice and support.

## Author Biography

*Zhenyi Liu received his MS in Electrical Engineering at Ulsan National Institute of Science and Technology, UNIST (2015) in Korea and is currently a PhD candidate in Automotive Engineering at Jilin University(2016-present), China and a Visiting Student Researcher at Stanford University(2017-present). His research interests focus on machine perception systems for autonomous vehicles such as cameras and lidar.*

*Minghao Shen is currently pursuing Bachelor degree in vehicle engineering with Jilin University, China(2015-now). He was research intern in Brian Wandell's lab at Stanford University(July 2018- September 2018). His research interests include applying machine learning methods to perception and motion planning for autonomous driving.*

*Shuangting Liu is a senior student in Automation Science and Electrical Engineering Department at Beihang University and will receive her B.S. degree in June, 2019. She was once a research intern at Vision Imaging Science and Technology Lab at Stanford University in 2018, supervised by Prof. Brian Wandell. Now she works as an intern at Microsoft Research Asia. Her research interest lies in computer vision, machine learning and computer graphics.*

*Jiaqi Zhang is an undergraduate student major in Pattern Recognition and Intelligent System in Automation Science and Electrical Engineering department at Beihang University (2015-now) and will receive his B.S. degree in June, 2019. He once was a research intern at Vision Imaging Science and Technology Lab at Stanford University, 2018, supervised by Prof. Brian Wandell. His research interests lies at computer vision and computer graphics, especially for applications in autonomous driving.*

*Henryk Blasinski received a MS degree in Telecommunications and Computer Science from the Lodz University of Technology, Lodz, Poland (2008), and the Diplome d'Ingeneiur degree from the Institute Superieur d'Electronique de Paris, France (2009). He graduated with a PhD in Electrical Engineering from Stanford University (2018). Henryk's research interests lie at the intersection of imaging, computer vision and machine learning.*

*Trisha Lian received her BS in Biomedical Engineering from Duke University (2014) and is currently a PhD student in Electrical Engineering at Stanford University. Her work has focused on the development of simulation tools for novel camera systems, as well as simulation of the human visual system*

*Brian A. Wandell is the first Isaac and Madeline Stein Family Professor. He joined the Stanford Psychology faculty in 1979 and is a member, by courtesy, of Electrical Engineering, Ophthalmology, and the Graduate School of Education. He is Director of Stanford's Center for Cognitive and Neurobiological Imaging and Deputy Director of Stanford's Neuroscience Institute. Wandell's research centers on vision science, spanning topics from visual disorders, reading development in children, to digital imaging devices and algorithms for both magnetic resonance imaging and digital imaging.*